\documentclass[letterpaper, 10 pt, conference]{ieeeconf}
\usepackage{amsmath,amsfonts}
\usepackage{array}
\usepackage{xcolor}
\definecolor{darkorange}{rgb}{0.8, 0.33, 0}
\definecolor{darkgreen}{rgb}{0, 0.5, 0}
\definecolor{lightblue}{rgb}{0.8, 0.9, 1.0}

\usepackage{enumerate}
\usepackage[utf8]{inputenc}
\usepackage[T1]{fontenc}
\usepackage{amsfonts}
\usepackage{amssymb}
\usepackage{tabularx}
\makeatletter
\let\NAT@parse\undefined
\makeatother
\usepackage{hyperref}
\hypersetup{
    colorlinks=true,
    linkcolor=blue,
    filecolor=magenta,      
    urlcolor=cyan,
    }
\IEEEoverridecommandlockouts 
\usepackage{algorithm}  
\usepackage{algpseudocode}  
\usepackage{amsmath}  
\usepackage{breqn}
\usepackage[caption=false]{subfig}
\usepackage{textcomp}
\usepackage{stfloats}
\usepackage{url}
\usepackage{verbatim}
\usepackage{graphicx}
\usepackage{cite}
\usepackage{color}
\usepackage{colortbl}
\let\labelindent\relax
\usepackage{enumitem}
\usepackage{multirow}
\usepackage{diagbox}
\usepackage{booktabs}
\usepackage{adjustbox}

\usepackage{siunitx}
\usepackage{rotating}

\title{\LARGE \bf Personalization in Human-Robot Interaction through \\ Preference-based Action Representation Learning}

\author{Ruiqi Wang$^{1\dag}$, Dezhong Zhao$^{1,2\dag}$, Dayoon Suh$^{1}$, Ziqin Yuan$^{1}$, Guohua Chen$^{2}$, and Byung-Cheol Min$^{1}$
\thanks{$^{1}$SMART Laboratory, Department of Computer and Information Technology, Purdue University, West Lafayette, IN, USA. {\tt\small{[wang5357, suh65, yuan460, minb]@purdue.edu}.}}
\thanks{$^{2}$College of Mechanical and Electrical Engineering, Beijing University of Chemical Technology (BUCT), Beijing, China. \tt\small{[DZ\_Zhao, chengh]@buct.edu.cn}.}
\thanks{$^\dag$These authors contributed equally.}
\thanks{The user study in this work has been reviewed by the Institutional Review Board (IRB) at BUCT, where it was conducted.}
}

\begin{document}

\setlength{\abovedisplayskip}{1pt} 
\setlength{\belowdisplayskip}{1pt} 

\maketitle

\begin{abstract}
Preference-based reinforcement learning (PbRL) has shown significant promise for personalization in human-robot interaction (HRI) by explicitly integrating human preferences into the robot learning process. However, existing practices often require training a personalized robot policy from scratch, resulting in inefficient use of human feedback. In this paper, we propose preference-based action representation learning (PbARL), an efficient fine-tuning method that decouples common task structure from preference by leveraging pre-trained robot policies. Instead of directly fine-tuning the pre-trained policy with human preference, PbARL uses it as a reference for an action representation learning task that maximizes the mutual information between the pre-trained source domain and the target user preference-aligned domain. This approach allows the robot to personalize its behaviors while preserving original task performance and eliminates the need for extensive prior information from the source domain, thereby enhancing efficiency and practicality in real-world HRI scenarios. Empirical results on the Assistive Gym benchmark and a real-world user study (N=8) demonstrate the benefits of our method compared to state-of-the-art approaches. Website at \url{https://sites.google.com/view/pbarl}.
\end{abstract}

\section{Introduction}
Propelled by advancements in machine learning, robots are increasingly engaging with humans in the daily life, performing varied roles from deliverers to assistants and companions \cite{gasteiger2023factors,wang2023initial}. Unlike industrial contexts, human-robot interaction (HRI) scenarios are remarkably nuanced and dynamic \cite{wang2024initial,semeraro2023human}. Humans, as individuals, often have distinct expectations when interacting with robots, even for identical tasks \cite{lee2012personalization}. Consequently, it is impractical to pre-train a robot policy to meet all potential requirements across diverse user preferences. Thus, robots should be capable of personalizing interactive behaviors to cater to individual user needs \cite{irfan2019personalization}.

Existing approaches to robot personalization generally fall into rule-based and learning-based categories \cite{hellou2021personalization}. Rule-based methods \cite{nikolaidis2017human,ikemoto2012physical,khoramshahi2019dynamical,peternel2016adaptation,john2022personalized,sung2009pimp,lee2012personalization,clabaugh2019long,schneider2021comparing,wu2023tidybot} tend to build user cognitive models that serve as the rule to guide the selection and adjustment of interaction behaviors. While straightforward, these methods are limited to pre-defined adaptation sets, which may not fully capture the complexity of human expectations and require significant effort to construct accurate user models.

On the other hand, learning-based approaches provide a more efficient and flexible alternative by enabling robots to learn from evolving interaction information streams. Within this category, an emerging direction is preference-based reinforcement learning (PbRL) \cite{akrour2012april,christiano2017deep,wang2022feedback,lee2021pebble,kumar2020discor,lee2021b,spencer2022expert,kupcsik2018learning,ritschel2019adaptive,wang2020teaching,hindemith2022interactive,zhang2023dual,liu2022task,wang2024prefclm,hejna2023few}. These methods aim to learn a preference-aligned reward model from continuous human comparative feedback during interactions, which then infers a personalized robot policy through RL. Compared to other learning-based approaches such as traditional RL \cite{khamassi2018robot,munguia2023affordance,del2022learning,ritschel2017real,jevtic2018robot,roveda2020model} and imitation learning \cite{de2022learning,mandlekar2020learning,qin2021dexmv,ramrakhya2022habitat,matsusaka2009health,sosa2022learning,rozo2016learning}, PbRL can avoid the intricate reward engineering to model subtle user preference, as well as the reliance on expert demonstrations, fostering more user-centred and natural personalization.

\begin{figure}[!t]
\centering
\includegraphics[width=\linewidth]{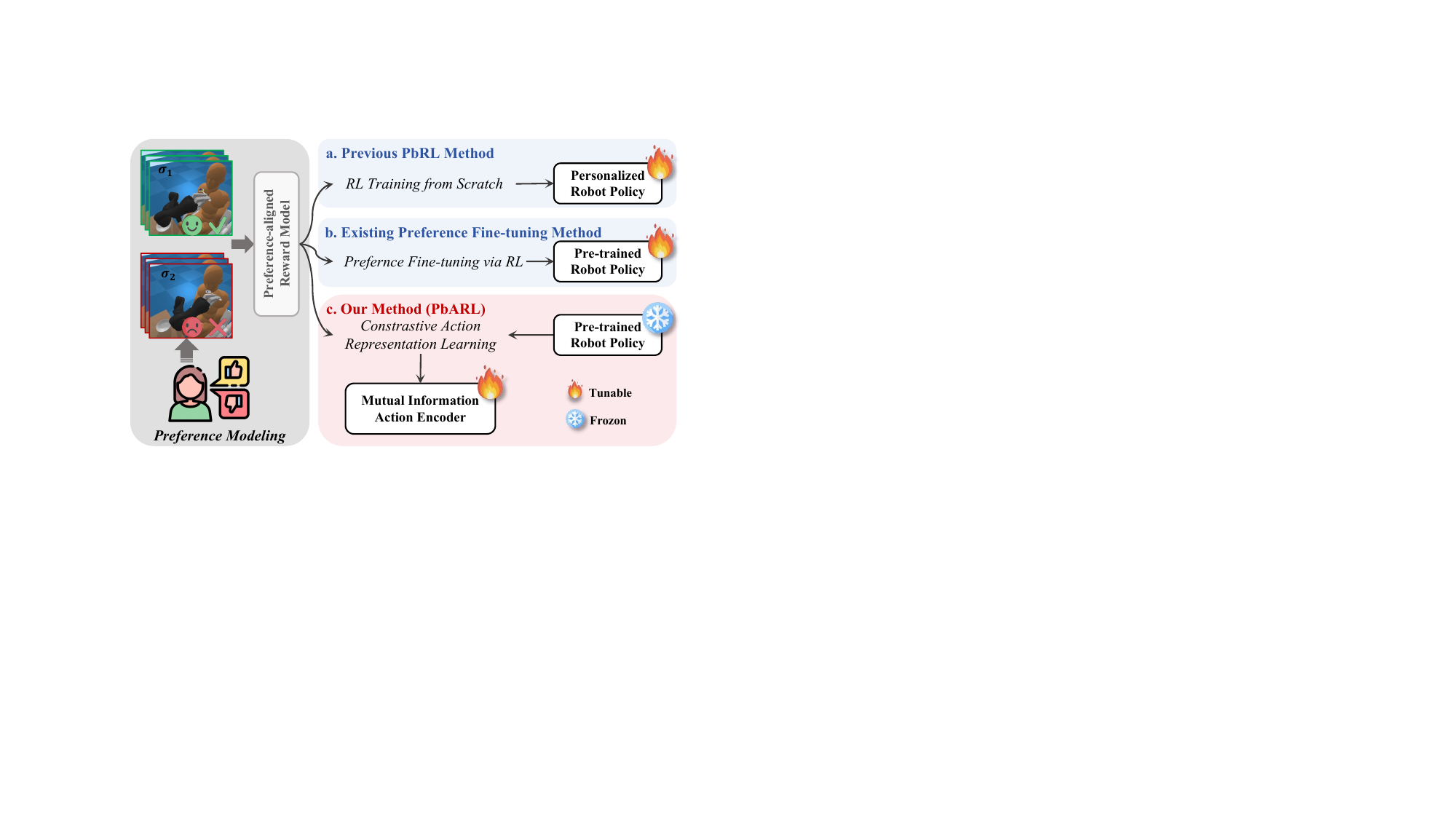}
\vspace{-20pt}
\caption{Comparison of our method with previous preference-based approaches for personalized adaptations. Unlike the common PbRL regime, which trains personalized policies from scratch, our method shifts toward fine-tuning to leverage human feedback more efficiently. Instead of using the preference-aligned reward model to directly adjust the pre-trained policy via RL, we employ it for an action representation task to train a mutual information encoder, preserving the pre-trained task performance while enhancing personalization.}
\vspace{-15pt}
\label{fig:comp}
\end{figure}

However, the predominant practice of current PbRL methods \cite{akrour2012april,christiano2017deep,wang2022feedback,lee2021pebble,kumar2020discor,lee2021b,spencer2022expert,kupcsik2018learning,ritschel2019adaptive,wang2020teaching,hindemith2022interactive,zhang2023dual,liu2022task,wang2024prefclm} is to train a personalized robot policy from scratch using user-specific preference data, placing them in a highly inefficient regime for feedback efficiency. Human feedback is needed not only to encode user expectations but also to cover basic task requirements, leading to an unnecessarily high demand for human input and thus restricting their real-world applicability. In contrast, fields that have been widely employed in the real world, such as computer vision and natural language processing, utilize a different regime involving transferring or fine-tuning pre-trained models to adapt to new task scenarios \cite{zhuang2020comprehensive,song2023comprehensive,julian2020never}. This regime seems promising to solve the above challenge. For example, a recent work \cite{hejna2023few} applied a meta-learning framework \cite{finn2017model} in PbRL to enable few-shot learning for new task demands.

Nevertheless, existing attempts based on conventional few-shot learning \cite{hejna2023few} and fine-tuning \cite{julian2020never,choi2020fast} may not meet the requirements of personalization in HRI, as it represents a different paradigm: in addition to enhancing performance in the target domain (personalization levels), the performance of the source domain (basic task performance) needs to be maintained \cite{hellou2021personalization}. When fine-tuning a pre-trained robot policy with such methods, where the target domain is the sole objective during adaptation, there is a risk of degrading the original task performance, which in turn could negatively impact user satisfaction. Additionally, the extra human efforts required for meta-policy pre-training \cite{hejna2023few} and the assumption of access to data \cite{julian2020never} or heavy task prior \cite{choi2020fast} in the source domain may not be feasible in realistic HRI scenarios.

To address these challenges, we propose \textbf{P}reference-\textbf{b}ased \textbf{A}ction \textbf{R}epresentation \textbf{L}earning (PbARL), an efficient fine-tuning method for personalization in HRI. Rather than using the preference-aligned reward model in PbRL as a supervised signal to directly fine-tune the pre-trained robot policy via RL, we utilize it as a guide for an action representation learning task. Our objective is to learn a latent action space that maximizes mutual information between the original domain of the pre-trained policy and the preference-aligned domain. This is achieved by training a mutual information action encoder, designed as a conditional variational autoencoder (cVAE) \cite{sohn2015learning} for its strong ability to learn structured and contextual latent spaces, with carefully designed loss functions that balance task performance preservation and personalization. By imposing only a soft constraint on the action space while keeping the pre-trained policy frozen, PbARL leverages the fine-tuning regime without significantly compromising basic task performance. Additionally, PbARL requires minimal information from the source domain, using transition tuples through policy evaluation. The main comparison with existing preference-based methods for personalization is illustrated in Fig.~\ref{fig:comp}. Our key contributions are highlighted as:
\begin{itemize}[leftmargin=*]
    \item We introduce PbARL, a fine-tuning method for personalization in HRI that shifts away from the training from scratch regime in PbRL, enhancing the efficiency of human feedback utilization.
    \item We propose to learn a latent action space that maximizes the mutual information between the pre-trained source domain and the target user preference domain without altering the pre-trained policy, enabling personalization while preserving original task performance.
    \item PbARL requires minimal prior knowledge from the source domain and can be seamlessly integrated with any pre-trained robot policy trained using standard RL algorithms, thus enhancing its practicality in real-world HRI scenarios.
    \item Through extensive experiments on the Assistive Gym benchmark \cite{erickson2020assistive} and a real-world user study (N=8), we demonstrate that PbARL can lead to greater user satisfaction in HRI, offering improved personalization levels and less degradation in task performance compared to existing state-of-the-art methods.
\end{itemize}

\section{Preliminary and Background}
\label{PF}
\subsection{Personalization in HRI and Preference-based RL}
\label{PbRL}
Personalization in HRI aims to tailor robot behaviors to meet the unique preferences and expectations of individual users \cite{irfan2019personalization}. Recently, preference-based RL has emerged as a promising direction due to its ability to directly incorporate user preferences into the learning process, eliminating the need for complex social reward function design and expensive expert demonstrations \cite{liu2022task}. 

The core objective of PbRL is to learn a preference-aligned reward model $\widehat{R}_\psi$, usually implemented as a neural network with parameters $\psi$, using a user preference dataset $\mathcal{D}_{\mathrm{pref}}=\left\{\left(\sigma_i^0, \sigma_i^1, \mathrm{y}_i\right)\right\}_{i=1}^{\left|\mathcal{D}_{\mathrm{pref}}\right|}$. Here, $\sigma$ represents a robot behavior trajectory, which is a sequence of state-action pairs across $T$ time steps, i.e., $\sigma = \left\{\left(\mathbf{s}_1, \mathbf{a}_1\right), \ldots,\left(\mathbf{s}_T, \mathbf{a}_T\right)\right\}$. The preference label $\mathrm{y}$ is determined as follows: $\mathrm{y} = 0$ if $\sigma^0$ is preferred over $\sigma^1$, $\mathrm{y} = 1$ if $\sigma^1$ is preferred over $\sigma^0$, and $\mathrm{y} = 0.5$ if both are equally preferred.

To learn such a reward model, a preference predictor based on Bradley-Terry model \cite{christiano2017deep} is utilized to calculate the preference probabilities as: 
\begin{equation}
\mathcal{P}_\psi\left[\sigma^1 \succ \sigma^0\right]=\frac{\exp \left(\sum_t \widehat{R}_\psi\left(\mathbf{s}_t^1, \mathbf{a}_t^1\right)\right)}{\sum_{j \in\{0,1\}} \exp \left(\sum_t \widehat{R}_\psi\left(\mathbf{s}_t^j, \mathbf{a}_t^j\right)\right)}
\label{BT}
\end{equation}
\noindent where $\mathcal{P}_\psi\left[\sigma^1 \succ \sigma^0\right]$ denotes the likelihood that the trajectory segment $\sigma^1$ is favored over $\sigma^0$.

Then the $\widehat{R}_\psi$ is learned by minimizing a cross-entropy loss between the preference labels $\mathrm{y}$ and the estimated preference probabilities as:
\begin{equation}
\begin{split}
\mathcal{L}_{\psi}=-\sum_{\left(\sigma^{0}, \sigma^{1}, \Lambda \right) \in \mathcal{D}_{\mathrm{pref}}} &(1-\mathrm{y}) \log \mathcal{P}_{\psi}\left[\sigma^{1} \succ \sigma^{0}\right]+ \\ &\mathrm{y} \log \mathcal{P}_{\psi}\left[\sigma^{0} \succ \sigma^{1}\right]
\end{split}
\label{loss}
\end{equation}

Despite significant advancements in preference modeling \cite{early2022non,kim2022preference,zhu2024decoding,kim2024guide}, which has been proven to align reward models with user preferences well, a key limitation hinders its utility in real-world situations. The prevalent PbRL regime \cite{akrour2012april,christiano2017deep,wang2022feedback,lee2021pebble,kumar2020discor,lee2021b,spencer2022expert,kupcsik2018learning,ritschel2019adaptive,wang2020teaching,hindemith2022interactive,zhao2024prefmmt,liu2022task,wang2024prefclm} necessitates training a personalized robot policy from scratch using the learned $\widehat{R}_\psi$ via RL, resulting in a highly inefficient utilization of human inputs. Specifically, since the distilled preference-aligned reward model serves as the sole reward signal for policy learning, human feedback is required not only to capture individual expectations but also to cover latent aspects of basic task requirements. This inefficiency is particularly pronounced in HRI scenarios, where the fundamental task structure often remains consistent across users, with personalization primarily affecting nuanced aspects of robot behaviors \cite{lee2012personalization}. 

Our work aims to bridge this gap by inducing a shift in the PbRL regime for personalized HRI, transitioning from learning from scratch to fine-tuning. By leveraging pre-trained robot policies that capture common task structures and basic interaction skills, we can allocate human feedback more efficiently to personalization aspects, minimizing the burden on human interactants and enabling more rapid adaptation.



\subsection{Transfer Learning and Fine-tuning}
Transfer learning, which aims to leverage knowledge gained in a source domain to address new task demands in a target domain, has emerged as a promising approach for fast adaptation \cite{zhuang2020comprehensive}. In fact, PbRL has been applied as an effective fine-tuning method to align pre-trained large language models (LLMs) with user preferences \cite{ouyang2022training,rafailov2024direct}. 

While it seems intuitive to apply this pattern to improve feedback efficiency of PbRL in HRI personalization, the scenario presents unique challenges. When adapting to user preferences in the target domain, the pre-trained task performance in the source domain must be maintained \cite{hellou2021personalization}. This contrasts with traditional fine-tuning, where the focus is primarily on the target domain \cite{zhuang2020comprehensive}. Additionally, adaptations in LLMs often involve merely adjusting response priorities \cite{ouyang2022training}, while personalization in HRI can lead to more nuanced changes in action selection and execution, which are more likely to conflict with the core pre-trained strategies. Our method addresses these challenges by learning a mutual information action space that balances the preservation of pre-trained task performance with personalized adaptations.

On the other hand, there have been efforts to incorporate transfer learning with robot learning \cite{zhu2023transfer}. For example, \cite{hejna2023few} proposes a meta-learning approach for PbRL, and \cite{julian2020never} utilizes data reuse to enable efficient robot policy fine-tuning. However, these methods either require extra human effort for meta-training or assume access to heavy task priors in source domain. In contrast, our method can be integrated with readily available pre-trained robot policies in RL without the need for expensive meta-training, and requires minimal information from the source domain, utilizing the state-action transition data obtained by testing the robot policy. 

\begin{figure*}[!t]
\centering
\includegraphics[width=0.95\linewidth]{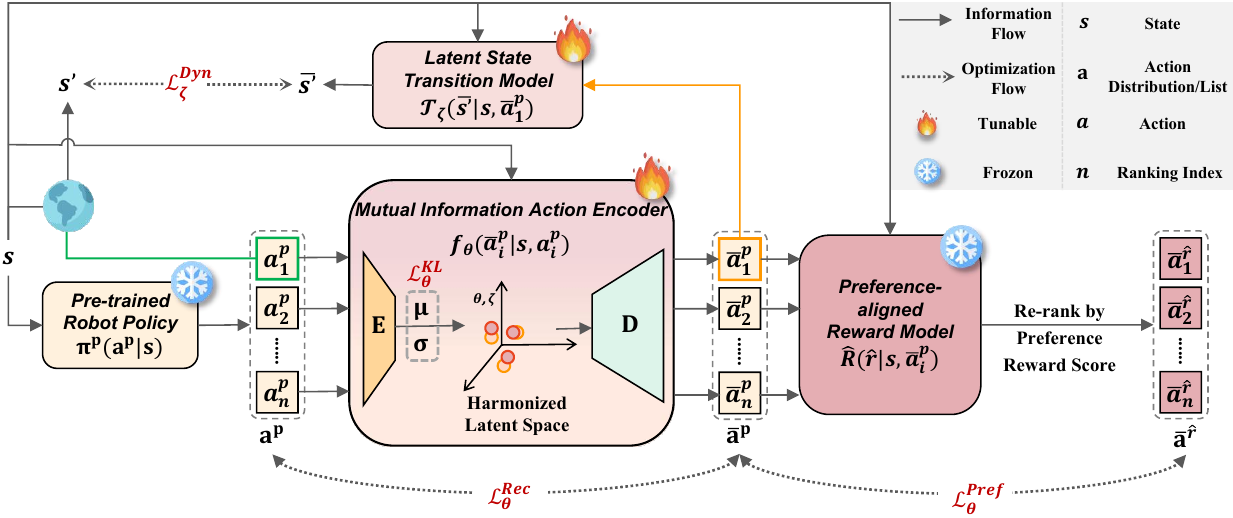}
\vspace{-10pt}
\caption{Overview of PbARL. We train PbARL using transition tuples: current state $s$, action distribution $\mathbf{a}$, and next state $s'$, collected by testing a pre-trained robot policy $\pi^p$ in the environment. The objective is to learn a harmonized latent action space within the mutual information state encoder $f_\theta$, implemented as a conditional VAE, by collectively optimizing three losses: a reconstruction loss $\mathcal{L}^{Rec}_\theta$, a preference loss $\mathcal{L}^{Pref}_\theta$ that reflects the consistency between the original action ranking list $\overline{\mathbf{a}}^p$ and the re-ranked action list $\overline{\mathbf{a}}^{\hat{r}}$ based on scores derived from the preference-aligned reward model $\hat{R}$, and a Kullback–Leibler (KL) loss $\mathcal{L}^{KL}_\theta$ to regularize the latent space in the VAE structure. To enhance controllability and scalability in the learned latent action space, we also conduct an auxiliary task to train a latent transition model $\mathcal{T}\varsigma$, optimized via a dynamic loss $\mathcal{L}^{Dyn}_\varsigma$.}
\vspace{-15pt}
\label{fig:framwork}
\end{figure*}

\subsection{Action Representation Learning in RL}
While state representation learning has traditionally been the main focus, the benefits of action representation learning are becoming increasingly recognized \cite{chandak2019learning,fujimoto2024sale,zheng2024texttt,van2020plannable}, especially for test-time policy adaptation \cite{losey2020controlling,allshire2021laser,hansenself}. For instance, \cite{losey2020controlling} constructs a latent state space that effectively bridges human intent inputs with robot manipulation actions of a pre-trained policy, and \cite{allshire2021laser} learns disentangled action representations to enhance sub-task performance within a task family. Building on similar principles, we aim to infer a latent action space that maximizes mutual information between the pre-trained source domain and the user preference-aligned domain.


\section{Methodology}
As illustrated in Fig. \ref{fig:framwork}, PbARL is implemented as an action representation learning task, where a mutual information state encoder is trained by jointly optimizing a reconstruction loss, a preference loss, a KL loss, and an auxiliary dynamic loss using state-action transition tuples collected from testing the pre-trained robot policy.

\subsection{Pre-trained Robot Policy and Preference Reward Model}
In HRI, while diverse and subtle user preferences are hard to predict, basic task requirements are often common and definable \cite{gasteiger2023factors}. In light of this, pre-trained robot policies are increasingly prevalent, with robots being equipped with basic interaction capabilities and task execution skills as part of their deployment in real-world settings \cite{radosavovic2023robot,radosavovic2023real,yang2024robot01,jing2023exploring}. Our PbARL leverages such a pre-trained robot policy $\pi^{p}(\mathbf{a}^{p}|s)$, which outputs a distribution over actions $\mathbf{a}^{p}$ in response to a state observation $s$. This policy can be trained with any standard RL to serve as a foundation for personalization, decoupling basic task performance from user-specific preference and thus enhancing human feedback efficiency.

Furthermore, PbARL utilizes a preference-aligned reward model $\hat{R}(\hat{r}|s,a)$, which is trained prior with standard preference modeling methods in PbRL as mentioned in Section \ref{PbRL}, to incorporate user preference awareness. With recent advancements in preference modeling \cite{early2022non,kim2022preference,zhu2024decoding,kim2024guide}, the learned reward model can closely reflect human preference patterns, assigning higher preference rewards $\hat{r}$ to state-action pairs that are more in line with user desires.

\subsection{Personalization as a Mutual Information Problem}
Let $\mathcal{M}_s$ denote the source domain associated with the pre-trained robot policy $\pi^{p}$ that captures default task performance and basic interaction skills. Let $\mathcal{M}_t$ represent the target domain, which is aligned with user-specific expectations as revealed by the preference reward model $\hat{R}$. The objective of personalization in PbARL is to learn a latent domain, $\mathcal{M}_m$, in the action space that maximizes the mutual information, $\mathcal{I}_m \cong \mathcal{I}_s \cap \mathcal{I}_t$, between the source and target domains.

To achieve this, we design an action representation learning task. The initial step is to collect a dataset consisting of multiple transitions $(s, \mathbf{a}^p, s')$ by rolling out the pre-trained policy. For each action distribution in the tuples, we sample an action list of $n$ actions, ranked by their probabilities under the distribution: $\mathbf{a}^p = \{a^p_1, \cdots, a^p_n\}$. Each action $a^p_i$ in the list, paired with the state $s$, is then passed through an action encoder $f_{\theta}(\overline{a}^p_i|s, a^p_i)$. The encoder is a cVAE, where reconstructing the action is conditioned on the state $s$ to ensure that the action representation is contextually grounded in the state information. The output of the encoder is a new action list, $\overline{\mathbf{a}}^p = \{\overline{a}^p_1, \cdots, \overline{a}^p_n\}$. We then pass each action in this list, along with the state $s$, to the pre-trained preference reward model $\hat{R}$, which re-ranks the actions based on the preference reward values $\hat{r}$. This results in a new preference-ranked list: $\overline{\mathbf{a}}^{\hat{r}} = \{\overline{a}^{\hat{r}}_1, \cdots, \overline{a}^{\hat{r}}_n\}$. 

Intuitively, the information $\mathcal{I}_s$ of the source domain can be measured by the similarity between actions at corresponding positions in $\mathbf{a}^p$ and $\overline{\mathbf{a}}^p$, and the information $\mathcal{I}_t$ of the target domain can be reflected by the consistency between the ranked lists $\overline{\mathbf{a}}^p$ and $\overline{\mathbf{a}}^{\hat{r}}$. While the mutual information $\mathcal{I}_m$ is hard to define and optimize directly, we can jointly optimize $\mathcal{I}_s$ and $\mathcal{I}_t$ to inter a suitable harmonized latent action space. Given this, we first design a reconstruction loss between $\mathbf{a}^p$ and $\overline{\mathbf{a}}^p$ by combining the conventional VAE reconstruction loss \cite{baldi2012autoencoders} with a hinge loss \cite{kipf2019contrastive} to avoid the negative distance from increasing indefinitely as seen in \cite{van2020plannable}, formulated as:
\begin{equation}
    \mathcal{L}^{\text{Rec}}_\theta  =\frac{1}{n}\sum_{i=1}^n\left(\left\|a_i^p-\bar{a}_i^p\right\|_2^2+\max \left(0, \epsilon-\left\|a_i^p-\bar{a}_i^p\right\|_2^2\right)\right)
\end{equation}
where $\epsilon$ is a scale parameter usually set to 1 \cite{van2020plannable,kipf2019contrastive}.

We further design a contrastive preference loss between $\overline{\mathbf{a}}^p$ and $\overline{\mathbf{a}}^{\hat{r}}$, based on the InfoNCE loss \cite{oord2018representation} as follows:
\begin{equation}
\mathcal{L}^{\text{Pref}}_{\theta} = -\frac{1}{n} \sum_{i=1}^n \log \left[\frac{\exp \left(\operatorname{sim}\left(\bar{a}_i^p, \bar{a}_i^{\hat{r}}\right) / \tau\right)}{\frac{1}{n-1}\sum\limits_{j=1,\, j \neq i}^n \exp \left(\operatorname{sim}\left(\bar{a}_i^p, \bar{a}_j^{\hat{r}}\right) / \tau\right)}\right]
\end{equation}
where, $\tau$ is a temperature parameter as in the standard InfoNCE loss, and the similarity between two actions is measured as:
\begin{equation}
\operatorname{sim}\left(\bar{a}_i^p, \bar{a}_j^{\hat{r}}\right) = -\left\|\bar{a}_i^p-\bar{a}_j^{\hat{r}}\right\|_2^2
\end{equation}

In this formulation, actions ranked at the same position in both lists are considered positive pairs, while all other combinations are treated as negative pairs. To address the potential imbalance caused by the negative pairs greatly outnumbering the positive pairs, we average the loss contribution of the negative pairs. 

Notably, unlike previous works that focus solely on the top-1 action in the loss functions \cite{losey2020controlling,allshire2021laser,van2020plannable}, we leverage the entire action list to provide a more nuanced and comprehensive shaping of the action distribution in the latent space. Additionally, we differentiate the formulation of the preference loss from that of the reconstruction loss because the reconstruction loss emphasizes similarity at the level of individual actions, while the preference loss prioritizes consistency in the ordering of actions between lists.

Furthermore, while we can maximize the mutual information through the two losses above, the learned action space may lack stability during execution due to insufficient latent controllability, consistency, and scaling compared to the original Markov decision process dynamics in the source domain, as shown in \cite{losey2020controlling}. To address this, following \cite{van2020plannable,allshire2021laser}, we also build an auxiliary task to optimize a latent state transition model $\mathcal{T}_\varsigma$ during training, which predicts the next state in the latent action space as:
\begin{equation}
    \overline{s'} = s + f_\varsigma(\bar{a}_1^p)
\end{equation}
where $\bar{a}_1^p$ denotes the executed action (i.e., the top-1 action in the $\overline{\mathbf{a}}^p$), and $f\varsigma$ is a 3-layer feedforward network.

The model is optimized via a dynamic loss $\mathcal{L}^{\text{Dyn}}_\varsigma$ as:
\begin{equation}
    \mathcal{L}^{Dyn}_\varsigma  =\left\|s'-\overline{s'}\right\|_2^2
\end{equation}

Following the $\beta$-VAE \cite{higgins2017beta} method, the final loss of PbARL is a weighted sum of all aforementioned losses plus the inherent KL divergence loss in VAE \cite{kingma2013auto}, denoted as:
\begin{equation}
\mathcal{L}_{\text{PbARL}} = \beta^{\text{Rec}} \mathcal{L}_\theta^{\text{Rec}} + \beta^{\text{Pref}} \mathcal{L}_\theta^{\text{Pref}} + \beta^{\text{KL}} \mathcal{L}_\theta^{\text{KL}} + \beta^{\text{Dyn}} \mathcal{L}_\varsigma^{\text{Dyn}}
\end{equation}

By collectively optimizing all loss components, we infer a harmonized latent action space embedded within the VAE structure and characterized by parameters $\theta$ and $\varsigma$. 

For deployment, similar to \cite{losey2020controlling}, we add the trained mutual information action encoder after the pre-trained policy, allowing the robot to transform actions into the learned latent action space, which are then reconstructed into the original space by the VAE decoder. This can lead to actions that are both task-effective and more closely aligned with user preferences. Although the learned latent state transition model is not used during deployment, it shapes the latent action space by enforcing consistency with the original MDP transitions during training, resulting in more structured and robust action representations for deployment.

\section{Experiments and Results}
\label{exp}
\subsection{Experiments on the Benchmark}
\subsubsection{Setups}
We first evaluated the proposed PbARL on the Assistive Gym benchmark \cite{erickson2020assistive}, focusing on three distinct assistive HRI tasks: i) \texttt{Feeding}: The robot arm, equipped with a spoon carrying small balls representing food, must deliver the food into the user's mouth without spilling; ii) \texttt{Drinking}: The robot grips a container filled with water, represented by pellets, and assists the user in drinking by guiding the container to the user’s mouth; and iii) \texttt{Itch Scratching}: The robot holds a scratching tool and moves it to a target location on the user's right arm, providing gentle scratching without applying excessive force. We used the Kinova Jaco assistive arm model as the robot platform for all tasks in the simulation.

The benchmark provides two types of rewards. The first is the robot task reward $r^T$ which is specific to each task and defines basic performance expectations. The second type is the human preference reward $r^H$, which is unified across different tasks and reflects common user preference items: 
\begin{equation}
r^H = \omega \odot \left[C_d, C_v, C_f, C_{hf}, C_{fd}, C_{fdv}\right]
\end{equation}
where $\omega$ denotes the weight vector for each preference item: i) $C_d(s)$: cost for long distance from the robot's end effector to the target assistance location; ii) $C_v$: cost for high robot end effector velocities; iii) $C_f$: cost for applying force away from the target assistance location; iv) $C_{hf}$: cost for applying high forces near the target; v) $C_{fd}$: cost for spilling food/water on the human; and vi) $C_{fdv}$: cost for food/water entering the mouth at high velocities.

Note that $C_{fd}$ and $C_{fdv}$ are only valid for \texttt{Feeding} and \texttt{Drinking} tasks. By adjusting the weight vector $\omega$, one can simulate different specific types of user preferences. In our experiments, we consider three distinct types of users:
\begin{itemize}[leftmargin=*]
\item Neutral User: $\omega = [1.0, 1.0, 1.0, 1.0, 1.0, 1.0]$. This user type maintains balanced preferences across all items.
\item Cautious User: $\omega = [1.0, 2.0, 1.5, 2.5, 3.0, 2.0]$. This user type emphasizes safety and comfort with higher weights on velocity control, force limitation, and spill prevention.
\item Impatient User: $\omega = [2.0, 0.5, 0.75, 0.5, 1.5, 0.5]$. This user type prioritizes quick task completion with higher weight on direct movements and lower weights on velocity and force constraints, while still maintaining spill prevention.
\end{itemize}

We used the provided task reward $r^T$ to pre-train the robot policies $\pi^{p}$ and employed the weighted human preference reward $r^H$ of each user type as a scripted teacher, conducting 5000 preference queries per task and user type, following the approach in \cite{lee2021b}. These queries were then used to train preference-aligned reward models $\hat{R}$ for each user type and task in offline mode, using a state-of-the-art preference modeling method \cite{kim2022preference}.

We compared our PbARL with three baselines: i) Pre-trained Policy: This represents the performance of the original policy trained on task-specific rewards via RL; ii) PbRL-HRI \cite{liu2022task}: A state-of-the-art preference-based RL method for personalization that designs sketchy task rewards, such as the distance from the spoon, container, or tool to the human mouth or target arm location, to incorporate basic task priors in PbRL, improving feedback efficiency; and iii) Preference Fine-tuning (PrefFT) \cite{ouyang2022training}: Inspired by practices in LLM fine-tuning, this approach uses the learned reward model $\hat{R}$ to fine-tune the pre-trained policy $\pi^{p}$ through RL re-training via SAC. We also conducted an ablation study on the \texttt{Feeding} task to investigate the effect of the number of actions in the action list, specifically $n = \{1, 10, 20\}$.

For evaluation metrics, we reported the success rate to reflect basic task performance and the average preference reward returns from $\hat{R}$ as a measure of personalization, tested on 500 unseen scenarios for each task and user type.

\subsubsection{Implementation Details}
We followed the suggested setup provided in the benchmark \cite{erickson2020assistive} to pre-train each robot policy using Soft Actor-Critic (SAC) \cite{haarnoja2018soft} for $1.6 \times 10^7$ steps. For the PbRL-HRI baseline, we adhered to the original procedure in \cite{liu2022task} for each task. For PrefFT, we fine-tuned the pre-trained policies with the learned preference reward models $\hat{R}$ using SAC for $1 \times 10^5$ steps for each task and each user type. For PbARL and all ablation variants, we set $\epsilon = 1$, and $\beta^{\text{Rec},\text{Pref},\text{KL},\text{Dyn}} = \{1, 1.5, 0.1, 1\}$ for \texttt{Feeding} and \texttt{Drinking}, and $\{1, 1, 0.1, 1\}$ for the more challenging \texttt{Scratching}. The sample size $n$ of each action list in the default PbARL was set to $10$. The action encoder is implemented as a classic 3-layer cVAE \cite{kingma2013auto}. We used Adam \cite{kingma2014adam} as the optimizer with an initial learning rate of 1e-4 for $1.5 \times 10^7$ steps. All experiments were conducted on a workstation equipped with three NVIDIA RTX 4090 GPUs

\begin{table}
\centering
\caption{Performance of each model across tasks and user types in terms of success rate (Succ\%) and mean preference reward returns (Rew) on 500 unseen test scenarios. The changes compared to the pre-trained model are shown in the $\Delta$ columns with arrows (↑ for improvement, ↓ for decrease). Best results (excluding pre-trained) are bolded.}
\label{tab:results}
\resizebox{1\linewidth}{!}{%
\begin{tabular}{@{}l l *{6}{S[table-format=2.0] S[table-format=2.0,table-column-width=1.8em] S[table-format=2.1] S[table-format=2.1,table-column-width=2.2em]}@{}}
\toprule
& & \multicolumn{4}{c}{Neutral User} & \multicolumn{4}{c}{Cautious User} & \multicolumn{4}{c}{Impatient User} \\
\cmidrule(lr){3-6} \cmidrule(lr){7-10} \cmidrule(lr){11-14}
Task & Method & {Succ} & {$\Delta$} & {Rew} & {$\Delta$} & {Succ} & {$\Delta$} & {Rew} & {$\Delta$} & {Succ} & {$\Delta$} & {Rew} & {$\Delta$} \\
\midrule
\multirow{6}{*}{Feeding} 
& Pre-trained & 92 & {--} & 11.2 & {--} & 90 & {--} & 12.3 & {--} & 89 & {--} & 15.6 & {--} \\
& PbRL-HRI & 41 & {\textcolor{darkorange}{↓51}} & 12.1 & {\textcolor{darkgreen}{↑0.9}} & 42 & {\textcolor{darkorange}{↓48}} & 12.8 & {\textcolor{darkgreen}{↑0.5}} & 39 & {\textcolor{darkorange}{↓50}} & 18.9 & {\textcolor{darkgreen}{↑3.3}} \\
& PrefFT & 73 & {\textcolor{darkorange}{↓19}} & 45.6 & {\textcolor{darkgreen}{↑34.4}} & 69 & {\textcolor{darkorange}{↓21}} & 51.2 & {\textcolor{darkgreen}{↑38.9}} & 75 & {\textcolor{darkorange}{↓14}} & 44.8 & {\textcolor{darkgreen}{↑29.2}} \\
& PbARL(n=1) & 80 & {\textcolor{darkorange}{↓12}} & 44.8 & {\textcolor{darkgreen}{↑33.6}} & 79 & {\textcolor{darkorange}{↓11}} & 48.1 & {\textcolor{darkgreen}{↑35.8}} & 78 & {\textcolor{darkorange}{↓11}} & 46.2 & {\textcolor{darkgreen}{↑30.6}} \\
& PbARL(n=20) & 84 & {\textcolor{darkorange}{↓8}} & \textbf{53.2} & {\textcolor{darkgreen}{\textbf{↑42.0}}} & 83 & {\textcolor{darkorange}{↓7}} & 52.1 & {\textcolor{darkgreen}{↑39.8}} & \textbf{85} & {\textcolor{darkorange}{\textbf{↓4}}} & 55.7 & {\textcolor{darkgreen}{↑40.1}} \\
& PbARL & \textbf{88} & {\textcolor{darkorange}{\textbf{↓4}}} & 50.7 & {\textcolor{darkgreen}{↑39.5}} & \textbf{83} & {\textcolor{darkorange}{\textbf{↓7}}} & \textbf{54.7} & {\textcolor{darkgreen}{\textbf{↑42.4}}} & 84 & {\textcolor{darkorange}{↓5}} & \textbf{56.1} & {\textcolor{darkgreen}{\textbf{↑40.5}}} \\
\midrule
\multirow{4}{*}{Drinking} 
& Pre-trained & 79 & {--} & 14.6 & {--} & 78 & {--} & 16.1 & {--} & 81 & {--} & 14.2 & {--} \\
& PbRL-HRI & 29 & {\textcolor{darkorange}{↓50}} & 16.3 & {\textcolor{darkgreen}{↑1.7}} & 35 & {\textcolor{darkorange}{↓43}} & 17.5 & {\textcolor{darkgreen}{↑1.4}} & 30 & {\textcolor{darkorange}{↓51}} & 14.7 & {\textcolor{darkgreen}{↑0.5}} \\
& PrefFT & 63 & {\textcolor{darkorange}{↓16}} & 41.2 & {\textcolor{darkgreen}{↑26.6}} & 61 & {\textcolor{darkorange}{↓17}} & 40.7 & {\textcolor{darkgreen}{↑24.6}} & 70 & {\textcolor{darkorange}{↓11}} & 36.9 & {\textcolor{darkgreen}{↑22.7}} \\
& PbARL & \textbf{74} & {\textcolor{darkorange}{\textbf{↓5}}} & \textbf{51.8} & {\textcolor{darkgreen}{\textbf{↑37.2}}} & \textbf{70} & {\textcolor{darkorange}{\textbf{↓8}}} & \textbf{48.6} & {\textcolor{darkgreen}{\textbf{↑32.5}}} & \textbf{79} & {\textcolor{darkorange}{\textbf{↓2}}} & \textbf{51.1} & {\textcolor{darkgreen}{\textbf{↑36.9}}} \\
\midrule
\multirow{4}{*}{Scratching} 
& Pre-trained & 71 & {--} & 10.9 & {--} & 74 & {--} & 11.8 & {--} & 74 & {--} & 10.2 & {--} \\
& PbRL-HRI & 23 & {\textcolor{darkorange}{↓48}} & 13.2 & {\textcolor{darkgreen}{↑2.3}} & 21 & {\textcolor{darkorange}{↓53}} & 14.5 & {\textcolor{darkgreen}{↑2.7}} & 24 & {\textcolor{darkorange}{↓50}} & 13.9 & {\textcolor{darkgreen}{↑3.7}} \\
& PrefFT & 55 & {\textcolor{darkorange}{↓16}} & \textbf{43.3} & {\textcolor{darkgreen}{↑32.4}} & 56 & {\textcolor{darkorange}{↓18}} & 46.3 & {\textcolor{darkgreen}{↑34.5}} & 52 & {\textcolor{darkorange}{↓22}} & 47.5 & {\textcolor{darkgreen}{↑37.3}} \\
& PbARL & \textbf{67} & {\textcolor{darkorange}{\textbf{↓4}}} & 40.6 & {\textcolor{darkgreen}{↑29.7}} & \textbf{69} & {\textcolor{darkorange}{\textbf{↓5}}} & \textbf{51.6} & {\textcolor{darkgreen}{\textbf{↑39.8}}} & \textbf{70} & {\textcolor{darkorange}{\textbf{↓4}}} & \textbf{51.1} & {\textcolor{darkgreen}{\textbf{↑40.9}}} \\
\bottomrule
\end{tabular}%
}\vspace{-20pt}
\end{table}

\subsubsection{Results and Analysis}
Table \ref{tab:results} presents the performance of PbARL compared to other baselines. A zoomed version can be found on our \href{https://sites.google.com/view/pbarl/home}{project website}. In general, we observe that, compared to other personalization methods, PbARL consistently achieves a superior balance between personalization and task performance, with much higher personalization levels, as indicated by preference reward returns, and without significantly compromising the original task success rate of the pre-trained policy. 

The PbRL-HRI baseline performs poorly compared to the original paper \cite{liu2022task}, showing largely reduced success rates and minimal improvement in preference returns. This is primarily due to the limited feedback queries (5,000) in our experiments, compared to 50,000 in the original study. This contrast highlights a key limitation of traditional PbRL approaches: they require extensive human feedback for satisfactory performance, which is often impractical in real-world HRI. The performance gap underscores the advantages of the fine-tuning paradigm. By leveraging a pre-trained policy, PbARL allocates feedback more efficiently to personalized requirements, resulting in better feedback utilization.

\begin{figure}[t]
\centering
\includegraphics[width=0.9\linewidth]{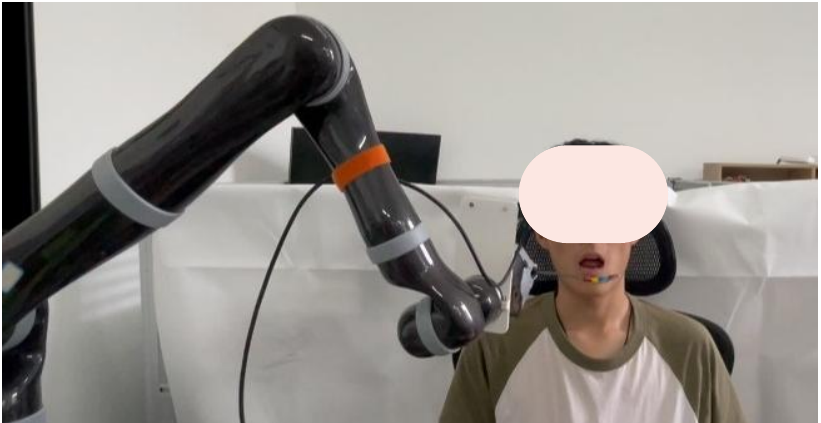}
\vspace{-10pt}
\caption{Depiction of the user study setup, where a seated participant interacts with a Jaco assistive robot arm for feeding using a spoon.}
\vspace{-20pt}
\label{fig:env}
\end{figure}

On the other hand, PrefFT shows sound improvements in personalization, even outperforming ours in 1 of 9 sub-settings. However, this is achieved at a high cost of task success. We attribute this to its optimization objective being solely focused on the target domain, which alters the pre-trained robot policy in a unidirectional manner. In contrast, PbARL maximizes the mutual information between the source and target domains through its learned harmonized latent action space, while keeping the pre-trained policy frozen. This enables PbARL to facilitate personalization without significantly compromising task performance acquired during pre-training, which in turn enhances personalization by maintaining task feasibility.

In addition, we observe that compared to $n$=1, the PbARL with $n$=10 and 20 performs better on \texttt{Feeding}, although $n$=20 does not consistently improve results across user types. This suggests that focusing on a broader action distribution is more efficient for constructing the latent action space compared to single action-level approaches in previous works \cite{losey2020controlling,allshire2021laser}. However, the optimal value of $n$ requires further investigation, which we leave for future research.

\subsection{Real-world User Study}
\subsubsection{Setups} To evaluate the personalization capabilities of our method in realistic HRI scenarios, we conducted a real-world user study. We selected the best-performing \texttt{Feeding} task for safety and practical considerations. The real-world setup, shown in Fig.~\ref{fig:env}, features a physical Kinova Jaco arm feeding a seated participant (similar to the simulation), a RealSense D435 camera, and a face landmark detection algorithm \cite{lugaresi2019mediapipe} to track the head and mouth positions of users. An emergency switch was also included to ensure user safety.

We recruited 8 participants (2 female, 6 male), with a mean age of 28.25 years (SD = 7.57). All participants signed informed consent forms. They were first introduced to the task and then interacted with the pre-trained robot policy, which was rolled out on a physical Kinova Jaco arm, completing 10 trials. In each trial, they interacted with a pair of trajectories and asked to provide their preference labels. Following the physical trials, participants were provided with videos of additional trajectory pairs, collecting another 190 preference labels. In total, 200 preference labels were collected per participant, which were used to train their individual preference reward models.

Subsequently, we used PbARL and the best-performing baseline, PrefFT, for personalization training in simulation, following the same settings mentioned above. Each participant interacted with the physical Jaco arm running each method along with the pre-trained policy for 3 trials each, presented in a random order. After each interaction, participants completed a questionnaire using a 1-5 Likert scale to rate their overall satisfaction and personalization levels of the robot behaviors, with 1 denoting not satisfied or personalized at all and 5 representing totally satisfied or personalized. Two-sample t-tests were conducted to determine the statistical significance of the performance differences between different methods.

\begin{figure}[t]
\centering
\includegraphics[width=\linewidth]{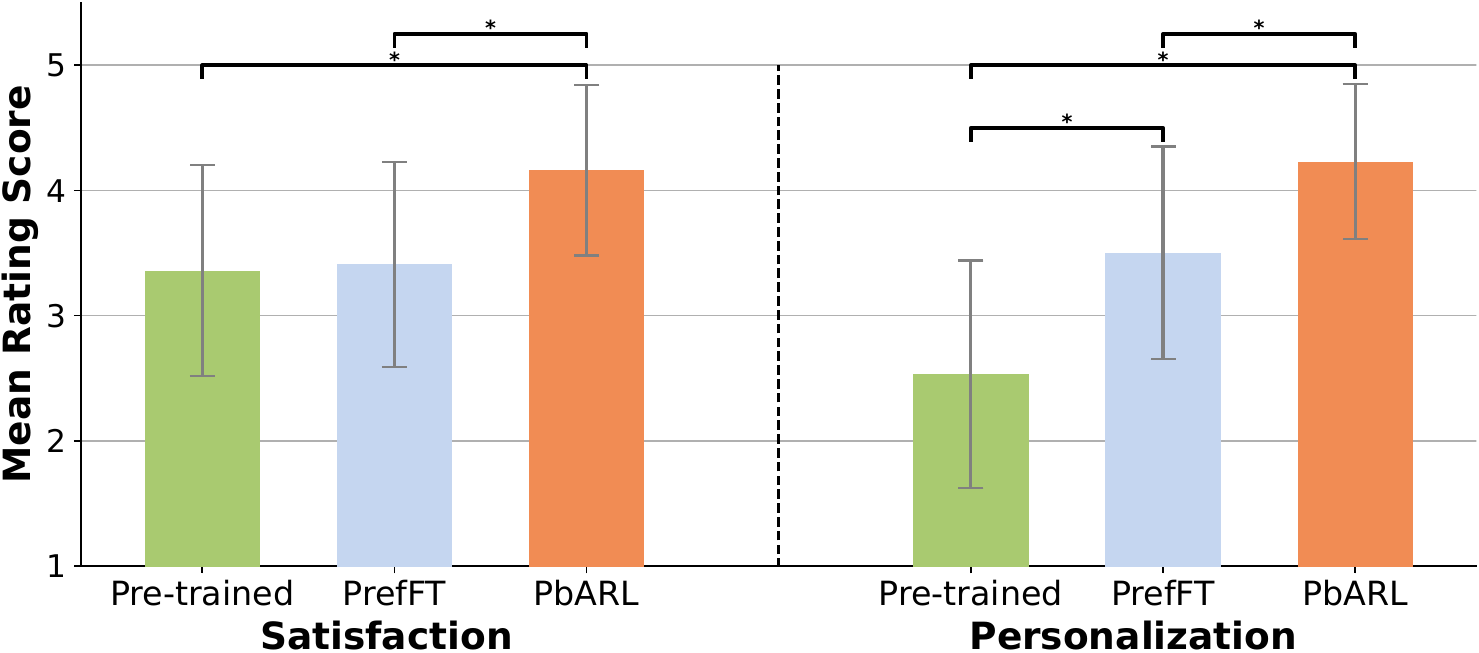}
\vspace{-20pt}
\caption{Mean rating scores of satisfaction and personalization levels for each method, as rated by participants. Results of two-sample t-tests are reported, with * indicating $p<.01$.}
\vspace{-20pt}
\label{fig:user}
\end{figure}

\subsubsection{Results and Analysis} Fig. \ref{fig:user} shows the the mean rating scores from the 8 participants and the t-test results. Our PbARL demonstrates statically significant improvements in both overall satisfaction and personalization levels compared to the pre-trained policy and PrefFT ($p<.01$), highlighting its effectiveness in real-world HRI scenarios. 

Interestingly, while PrefFT leads to higher personalization ratings ($p<.01$), its overall satisfaction scores are similar to those of the pre-trained policy ($p>.05$). We attribute this to the fact that, while PrefFT can personalize robot behaviors, e.g., slowing down for cautious users, it also causes performance degradation, such as not positioning the spoon close enough to the mouth. This trade-off results in improved perceived personalization but fails to enhance overall user satisfaction. These findings underscore the limitations of simply applying traditional fine-tuning that primarily focuses on the target domain in HRI personalization.

In contrast, PbARL formulates personalization as a mutual information maximization problem in a latent action space, without altering the pre-trained robot policy that contains essential task skills. This approach enables effective personalization without significantly compromising task performance, ultimately resulting in higher overall satisfaction. Demos of the user study are available on our \href{https://sites.google.com/view/pbarl/home}{project website}.

\section{Conclusion and Future work}
In this paper, we present a feedback-efficient fine-tuning method for personalization in HRI, named PbARL, leveraging pre-trained robot policies that capture foundational task skills. By learning a latent action space that maximizes mutual information between the pre-trained source domain and the target preference domain through a cVAE structure, it ensures effective personalization without significantly compromising task performance. Future work aims to explore extending PbARL for lifelong personalization to accommodate evolving and more complex human preferences in long-term HRI scenarios.

\typeout{}
\bibliography{main}
\bibliographystyle{IEEEtran}
\end{document}